\documentclass[10 pt, journal, twoside]{IEEEtran}
\usepackage{eurosym}
\usepackage{amsmath}
\usepackage{amsfonts}
\usepackage{fixmath}
\usepackage{graphicx}
\usepackage{enumitem}
\usepackage{color}
\usepackage{caption}
\usepackage{subcaption}

\IEEEoverridecommandlockouts

\newtheorem{lemma}{Lemma}
\newtheorem{proof}{Proof}
\input{tcilatex}
\begin{document}

\title{{\LARGE \textbf{A Recursive Lie-Group Formulation for the
Second-Order Time Derivatives of the Inverse Dynamics of Parallel Kinematic
Manipulators}}}
\author{Andreas M\"{u}ller$^{1}$, Shivesh Kumar$^{2}$, Thomas Kordik$^{1}$ 
\thanks{Manuscript received: November 30, 2022; Revised March 2, 2023; Accepted March 28, 2023.
This paper was recommended for publication by Editor Lucia Pallottino upon evaluation of the Associate Editor and Reviewers' comments. Support by LCM K2 Center for Symbiotic Mechatronics within the Austrian COMET-K2 program is acknowledged.}
\thanks{$^{1}$Institute of Robotics, Johannes Kepler University, 4040 Linz, Austria; $^{2}$Robotics Innovation Center, DFKI GmbH, 28359 Bremen, Germany} 
\thanks{Digital Object Identifier (DOI): see top of this page.}
}
\maketitle

\begin{abstract}
Series elastic actuators (SEA) were introduced for serial robotic arms.
Their model-based trajectory tracking control requires the second time
derivatives of the inverse dynamics solution, for which algorithms were
proposed. Trajectory control of parallel kinematics manipulators (PKM)
equipped with SEAs has not yet been pursued. Key element for this is the
computationally efficient evaluation of the second time derivative of the
inverse dynamics solution. 
This has not been presented in the literature, and is addressed in the
present paper for the first time. 
The special topology of PKM is exploited reusing the recursive algorithms
for evaluating the inverse dynamics of serial robots. A Lie group
formulation is used and all relations are derived within this framework.
Numerical results are presented for a 6-DOF Gough-Stewart platform (as part
of an exoskeleton), and for a planar PKM when a flatness-based control
scheme is applied.

\end{abstract}


\setcounter{topnumber}{3} \setcounter{bottomnumber}{2} \renewcommand{%
\textfraction}{0.00001}


\setcounter{topnumber}{3} \setcounter{bottomnumber}{2} \renewcommand{%
\textfraction}{0.00001}


\section{Introduction}

Series elastic actuators (SEA) were introduced as actuation concept for
serial kinematic robotic arms \cite{Pratt2002} as a means to provide
inherent compliance (a key characteristics of collaborative robots
(cobots)). Lightweight arms were proposed in order to reduce the moving
mass. Along this line, parallel kinematics manipulators (PKM) equipped with
SEA possess lower moving mass, and thus lower reflected inertia at the
end-effector (EE), compared to serial robots, which serves as criteria for
safety assessment \cite{Worsnopp2006,Huck2021,Kirschner2021}. They would
hence be perfectly suited as inherently compliant agile robotic manipulators
and cobots. Yet, SEA-actuated PKM (SEA-PKM) have exclusively been used as
force-controlled support devices \cite{Sergi2013,Erdogan2017,Lee2022}.
\mbox{The dynamics of SEA-PKM is expressed as}
\begin{align}
\mathbf{M}_{\mathrm{t}}\dot{\mathbf{V}}_{\mathrm{t}}+\mathbf{C}_{\mathrm{t}}%
\mathbf{V}_{\mathrm{t}}+\mathbf{W}_{\mathrm{t}}^{\mathrm{grav}}& =\mathbf{J}%
_{\mathrm{IK}}^{T}\mathbf{u}  \label{EOMTask1} \\
\mathbf{M}_{\mathrm{m}}\ddot{\mathbf{q}}_{\mathrm{m}}& =\pmb{\tau}-\mathbf{u}
\label{EOM2}
\end{align}%
where the equations of motion (EOM) (\ref{EOMTask1}) govern the PKM dynamics
in task space (see Sec. \ref{secInvDyn}), and (\ref{EOM2}) the actuator
dynamics, with $\mathbf{M}_{\mathrm{m}}=\mathrm{diag}(m_{1},\ldots m_{n_{%
\mathrm{a}}})$ defined by the reduced inertia moment $m_{i}$ of the $i$th
drive unit, and the vector $\pmb{\tau}$ of actuator torques/forces. Both are
coupled via the elastic forces $\mathbf{u}=\mathbf{K}(\mathbf{q}_{\mathrm{m}%
}-{\mathbold{\vartheta}}_{\mathrm{a}})$, where $\mathbf{K}=\text{diag}%
(k_{1},\ldots k_{n_{\mathrm{a}}})$ describes the SEA compliance, with
stiffness coefficient $k_{i}$ associated to drive $i$. Here ${%
\mathbold{\vartheta}}_{\mathrm{a}}$ denotes the coordinate vector with the $%
n_{\mathrm{a}}$ actuated joints, $\mathbf{q}_{\mathrm{m}}$ the vector of $n_{%
\mathrm{a}}$ motor coordinates, and $\mathbf{V}_{\mathrm{t}}$ the task space
velocity of the platform. The PKM mechanism is actuated by the elastic
forces $\mathbf{u}$, which are transformed to task space with the inverse
kinematics Jacobian $\mathbf{J}_{\mathrm{IK}}$ (i.e. $\dot{{%
\mathbold{\vartheta}}}_{\mathrm{a}}=\mathbf{J}_{\mathrm{IK}}\mathbf{V}_{%
\mathrm{t}}$).

\textbf{Problem:} Position and trajectory tracking control of SEA-driven
robots necessitate model-based feedforward control. Flatness-based exact
feedback linearization control methods are well established for SEA-actuated
serial robotic arms \cite%
{deLuca1998,GattringerMUBO2014,PalliMelchiorriDeLuca2008}. Since the model (%
\ref{EOMTask1}),(\ref{EOM2}) is formally identical to that of SEA-actuated
serial robots (except the IK Jacobian), they can be directly adopted to
SEA-PKM. The EE (task space) motion is used as flat output, and it is known
that the vector relative degree of this control system with input $\pmb{\tau}
$ is $\{4,\ldots ,4\}$, and $\{3,\ldots ,3\}$ if damping is included in (\ref%
{EOM2}). That is, the PKM state and the input $\pmb{\tau}$ can be expressed
in terms of EE pose, task space velocity $\mathbf{V}_{\mathrm{t}}$, and its
time derivatives $\dot{\mathbf{V}}_{\mathrm{t}},\ddot{\mathbf{V}}_{\mathrm{t}%
},\dddot{\mathbf{V}}_{\mathrm{t}}$. To this end, (\ref{EOMTask1}) is solved
for $\mathbf{q}_{\mathrm{m}}$. Substituting this solution, and its second
time derivative in (\ref{EOM2}) yields $\pmb{\tau}$, which serves as
feed-forward control. The crucial aspect of the flatness-based control is
that it \emph{involves the first and second time derivative of the EOM} (\ref%
{EOMTask1}). For serial robotic arms, recursive second-order inverse
dynamics $O\left( n\right) $ algorithms were proposed \cite%
{BuondonnaDeLuca2016,Guarino2009} using classical vector formulations of
rigid body kinematics. Recursive Lie group formulations were proposed in 
\cite{ICRA2017,RAL2020} employing compact expressions for rigid body twists
complementing the inverse dynamics algorithms in \cite%
{ParkBobrowPloen1995,LynchPark2017}, which can be seen as generalization of
the spatial vector algebra \cite{Rodriguez1991,JainBook}. Also a closed form
Lie group formulation was reported in \cite{MUBO2021}. All these approaches
for serial kinematics robots are direct extensions of recursive inverse
dynamics formulations. In contrast, the derivatives of the inverse dynamics
solution for PKM is more involved due to the presence of loop constraints.
The latter can be resolved and incorporated in the EOM in various different
ways, and the modeling approach dictates the complexity of the higher-order
inverse dynamics algorithm, which is thus crucial to for development of
SEA-PKM into robotic manipulators. The best suited modeling method is the
one proposed for non-redundant fully parallel PKM with simple limbs reported
in \cite{AngelesBook,BriotKhalilBook} and \cite{SahaSchiehlen2001}. In this
method, each limb is regarded as a serial chain, and their motion is
expressed in terms of the platform motion by means of the inverse kinematics
solution of each limb. For each limb, this resembles a task space
formulation of serial robots. This was developed into a \emph{task space }%
formulation for general redundant PKM with simple limbs in \cite%
{MuellerAMR2020} and with complex limbs in \cite{MMT2022,MuellerRAS2022}
using a Lie group framework (although the basic concept does not depend on
it). The important implication is that the higher-order inverse dynamics
problem boils down to merging the higher-order inverse dynamics of the
individual limbs with the higher-order inverse kinematics of
the PKM.

\textbf{Contribution:} In this paper, for the first time in the literature,
a second-order inverse dynamics algorithm for computing the second time
derivative of (\ref{EOMTask1}) for non-redundant PKM with simple limbs is
proposed. It builds upon the dynamics formulation introduced in \cite%
{MuellerAMR2020,MMT2022} and the recursive inverse dynamics algorithms for
serial robots introduced in \cite{ICRA2017,RAL2020} using a Lie group
formulation. For an accessible introduction to the general Lie group
formulation, the reader is referred to \cite{Murray,LynchPark2017}, while a
summary of the particular formulation used in this paper can be found in 
\cite{MUBOScrews1,MUBOScrews2}. Due to space limitation, the algorithm is
presented for kinematically non-redundant PKM only, while it is applicable
to general fully parallel PKM. The actual (flatness-based) control using the
presented second-order inverse dynamics solution will be topic of a
forthcoming publication. This paper provides the algorithmic foundation, as 
\cite{deLuca1998}-\cite{RAL2020} do for serial robots.

\textbf{Organization:} Sec. \ref{secKinematics} recalls the kinematic
modeling of PKM emphasizing that a single limb can be treated as a serial
kinematic chain. The closed form inverse dynamics formulation is summarized
and expressed in a form suited for computing time derivatives in Sec. \ref%
{secInvDyn}. The new algorithm for computing the 4th time derivatives of the
inverse and forward kinematics is presented in Sec. \ref{secFKIK}, which is
then used for the second time derivative of the inverse dynamics in Sec. \ref%
{secHighInvdyn}. Numerical results are presented in Sec. \ref{secExamples}
for a 6U\underline{P}S Gough-Stewart platform (GSP) and  a planar 2-DOF 3%
\underline{R}RR PKM. Outlook \mbox{and suggestions for future research are given
in Sec. \ref{secConclusion}.}

\section{Kinematics%
\label{secKinematics}%
}

\subsection{Kinematic topology}

\begin{figure}[t]
\centerline{
\hfill
a)\includegraphics[height=4.2cm]{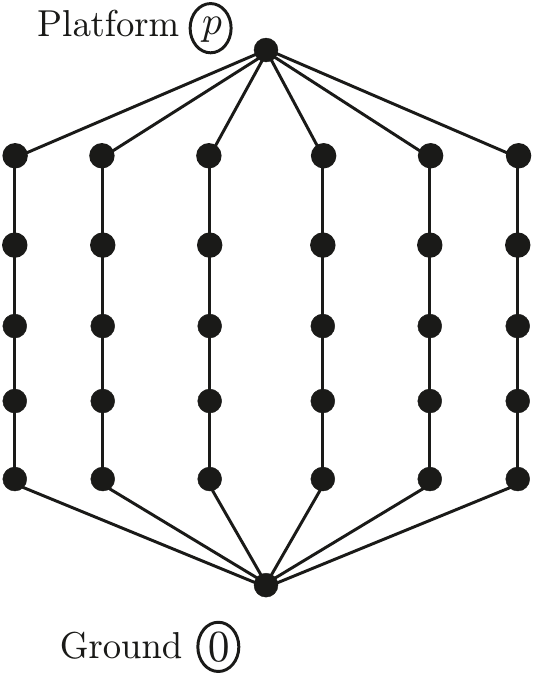}
\hfill
b)\includegraphics[height=4.1cm]{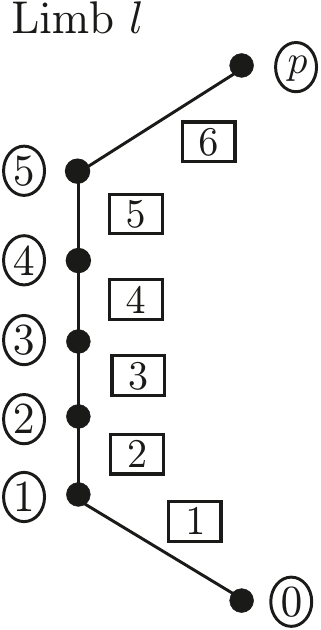}
\hfill}
\caption{a) Topological graph of a Gough-Stewart platform. b) Sub-graph
corresponding to one limb including platform.}
\label{figHexapod}
\end{figure}

A fully parallel PKM consists of a moving platform connected to the fixed
platform (ground) by $L$ limbs. Each limb is (in this paper) a serial
kinematic chain (therefore called simple). A typical example is the GSP
whose topological graph is shown in Fig. \ref{figHexapod}a). Following the
common convention, technical joints are modeled as combination of 1-DOF
joints, so that each limb $l=1,\ldots ,L$ comprises $N_{l}=6$ joints
(edges). Topologically, the PKM consists of $L$ congruent sub-graphs
connected to the platform, one is shown in Fig. \ref{figHexapod}b). This
gives rise to a tailored kinematics modeling, where the kinematics of each
sub-graph is described in terms of the platform motion. 

\subsection{Forward and inverse kinematics of limbs}

Denote with $%
\mathbold{\vartheta}%
_{\left( l\right) }\in {\mathbb{V}}^{N_{l}}$ the vector of $N_{l}$ joint
coordinates of limb $l$ when connected to the platform. The velocity
'forward kinematics' of limb $l$ gives the platform twist $\mathbf{V}_{%
\mathrm{p}}$ in terms of the $6\times N_{l}$ \emph{forward kinematics
Jacobian of limb} $l$ 
\vspace{-1ex}%
\begin{equation}
\mathbf{V}_{\mathrm{p}}=\mathbf{J}_{\mathrm{p}\left( l\right) }\dot{%
\mathbold{\vartheta}%
}_{\left( l\right) }.  \label{FWKin}
\end{equation}%
This relates platform twist and joint rates $\dot{%
\mathbold{\vartheta}%
}_{\left( l\right) }$ when the platform is connected only to the separated
limb $l$.

The platform of the PKM, i.e. when connected to all limbs, has DOF $\delta _{%
\mathrm{p}}\leq N_{l}$, and only $\delta _{\mathrm{p}}$ components of $%
\mathbf{V}_{\mathrm{p}}$ are independent. The \emph{task space velocity}
vector $\mathbf{V}_{\mathrm{t}}$ is introduced accordingly comprising the $%
\delta _{\mathrm{p}}$ relevant components of the platform twist $\mathbf{V}_{%
\mathrm{p}}$. This is formally expressed as 
\vspace{-1ex}%
\begin{equation}
\mathbf{V}_{\mathrm{p}}=\mathbf{P}_{\mathrm{p}}\mathbf{V}_{\mathrm{t}}
\end{equation}%
with a unimodular $6\times \delta _{\mathrm{p}}$ velocity distribution
matrix $\mathbf{P}_{\mathrm{p}}$, which assigns the $\delta _{\mathrm{p}}$
components of the task space velocity to the components of the platform
twist. This relates platform twist and task space velocity. The latter is
used in the task space formulation of EOM. The intermediate step via the
platform twist is crucial to account for general PKM.

Consider the platform when only connected to limb $l$. The platform has DOF $%
\delta _{\mathrm{p}\left( l\right) }\geq \delta _{\mathrm{p}}$, which is the
generic rank of $\mathbf{J}_{\mathrm{p}\left( l\right) }$, whereas the
serial chain has DOF $N_{l}\geq \delta _{\mathrm{p}\left( l\right) }$. If $%
\delta _{\mathrm{p}\left( l\right) }=\delta _{\mathrm{p}}$, the PKM is
called \emph{equimobile} \cite{MuellerAMR2020,MMT2022}, i.e. there is a $%
\delta _{\mathrm{p}\left( l\right) }\times N_{l}$ submatrix $\mathbf{J}_{%
\mathrm{t}\left( l\right) }$ of $\mathbf{J}_{\mathrm{p}\left( l\right) }$ so
that $\mathbf{V}_{\mathrm{t}}=\mathbf{J}_{\mathrm{t}\left( l\right) }\dot{%
\mathbold{\vartheta}%
}_{\left( l\right) }$. For a non-equimobile PKM ($\delta _{\mathrm{p}\left(
l\right) }>\delta _{\mathrm{p}}$, i.e. the platform has a different mobility
when connected to the kinematic chain of a single limb and when connected to
all limbs), only $\delta _{\mathrm{p}\left( l\right) }$ rows correspond to
the task space velocity, while the remaining $\delta _{\mathrm{p}\left(
l\right) }-\delta _{\mathrm{p}}$ rows represent constraints on the platform
motion. This is expressed with help of a $\delta _{\mathrm{p}\left( l\right)
}\times \delta _{\mathrm{p}}$ velocity distribution matrix $\mathbf{D}_{%
\mathrm{t}\left( l\right) }$ so that%
\vspace{-1ex}%

\begin{equation}
\mathbf{D}_{\mathrm{t}\left( l\right) }\mathbf{V}_{\mathrm{t}}=\mathbf{J}_{%
\mathrm{t}\left( l\right) }\dot{%
\mathbold{\vartheta}%
}_{\left( l\right) }.  \label{VpD}
\end{equation}%
The $\delta _{\mathrm{p}\left( l\right) }\times N_{l}$ matrix $\mathbf{J}_{%
\mathrm{t}\left( l\right) }$ is the \emph{task space Jacobian of limb }$l$,
formally defined as $\mathbf{J}_{\mathrm{t}\left( l\right) }:=\mathbf{P}_{%
\mathrm{t}\left( l\right) }\mathbf{J}_{\mathrm{p}\left( l\right) }$, with $%
\delta _{\mathrm{p}\left( l\right) }\times 6$ selection matrix $\mathbf{P}_{%
\mathrm{t}\left( l\right) }$. The latter simply extracts the relevant rows
from the forward kinematics Jacobian. Throughout the paper, it is assumed
that the PKM is kinematically non-redundant, i.e. $\delta _{\mathrm{p}\left(
l\right) }=N_{l}$, and $\mathbf{J}_{\mathrm{t}\left( l\right) }$ is a full
rank $N_{l}\times N_{l}$ matrix, implying that the mechanism DOF $\delta $
is equal to $\delta _{\mathrm{p}}$.

Introducing the \emph{inverse kinematics Jacobian of limb }$l$, $\mathbf{F}%
_{\left( l\right) }:=\mathbf{J}_{\mathrm{t}\left( l\right) }^{-1}\mathbf{D}_{%
\mathrm{t}\left( l\right) }$, the solution of the inverse kinematics problem
at velocity and acceleration level is, respectively,%
\begin{align}
\dot{%
\mathbold{\vartheta}%
}_{\left( l\right) }& =\mathbf{F}_{\left( l\right) }\mathbf{V}_{\mathrm{t}%
},\ \ \ \mathrm{with\ \ }\mathbf{F}_{\left( l\right) }:=\mathbf{J}_{\mathrm{t%
}\left( l\right) }^{-1}\mathbf{D}_{\mathrm{t}\left( l\right) }
\label{Ftheta1} \\
\ddot{%
\mathbold{\vartheta}%
}_{\left( l\right) }& =\mathbf{F}_{\left( l\right) }\dot{\mathbf{V}}_{%
\mathrm{t}}+\dot{\mathbf{F}}_{\mathrm{t}\left( l\right) }\mathbf{V}_{\mathrm{%
t}}.  \label{Ftheta2}
\end{align}

\subsection{Kinematics of associated tree-topology system}

\begin{figure}[t]
\centerline{a)\includegraphics[height=4cm]{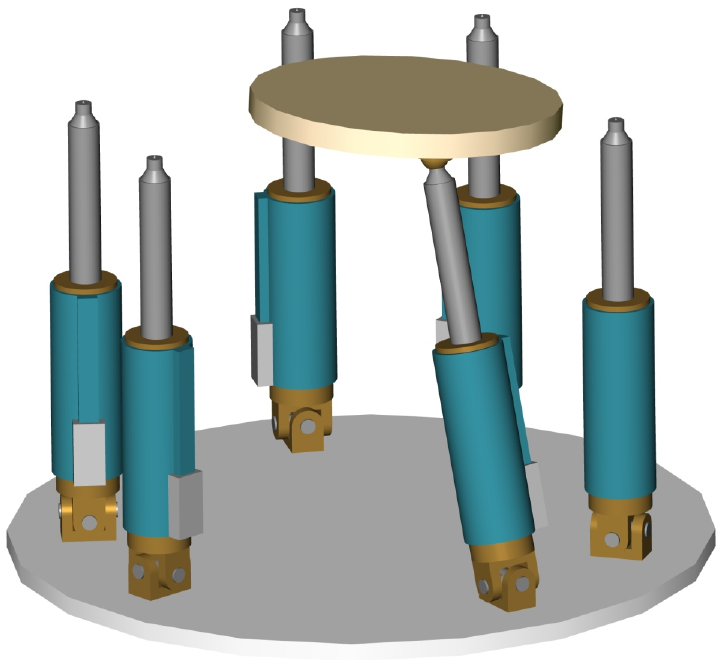}~~
b)\includegraphics[height=3.7cm]{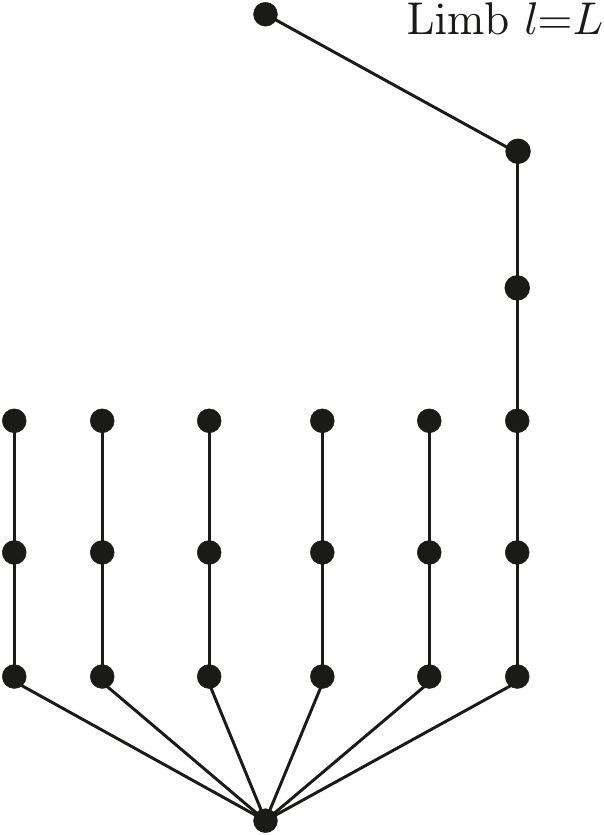}
}
\caption{a) Tree-topology model of Gough-Stewart PKM. Platform connected to
one limb.\thinspace \mbox{b) Corresponding spanning tree.}}
\label{figTreeHexapod}
\end{figure}
A tree-topology system is introduced by eliminating cut-joints, which is a
standard approach in multibody dynamics \cite%
{NikraveshBook1988,WittenburgBook,JainBook}. A fully parallel PKM possesses $%
L-1$ fundamental cycles. Taking into account the special topology of PKM,
the $L-1$ joints connecting the platform to the respective limb are selected
as cut-joints. That is, a spanning tree is introduced so that the platform
is the leaf of one serial chain, i.e. the platform remains attached to one
limb while it is disconnected from the remaining $L-1$ limbs. W.l.o.g.,
limbs $l=1,\ldots ,L-1$ are cut off from the platform. In the tree-topology
system, the platform is then connected to limb $L$, shown in Fig. \ref%
{figTreeHexapod} for the 6-DOF GSP.

Denote with $\bar{%
\mathbold{\vartheta}%
}_{\left( l\right) }\in {\mathbb{V}}^{n_{l}}$ the $n_{l}$ tree-joint
coordinates of limb $l=1,\ldots ,L-1$ when disconnected from the platform
(this is $%
\mathbold{\vartheta}%
_{\left( l\right) }$ with variables of the cut joint connecting to the
platform removed), and with $\bar{\mathbf{F}}_{\left( l\right) }$ the $%
n_{l}\times \delta _{\mathrm{p}}$ submatrix of $\mathbf{J}_{\mathrm{t}\left(
l\right) }^{-1}$ obtained by eliminating the rows corresponding to the
variables of the cut-joint connecting the limb to the platform. Then $\dot{%
\bar{%
\mathbold{\vartheta}%
}}_{\left( l\right) }=\bar{\mathbf{F}}_{\left( l\right) }\mathbf{V}_{\mathrm{%
t}}$. In case of the GSP, the 3-DOF spherical joints are cut, and the
remaining $L-1$ chains have $n_{l}=3$ joint coordinates. Limb $L$ does not
contain cut-joints, and comprises the platform in the tree-topology, so that 
$\bar{%
\mathbold{\vartheta}%
}_{\left( L\right) }=%
\mathbold{\vartheta}%
_{\left( L\right) }$ are the corresponding $N_{L}$ tree-joint coordinates.

Summarizing the tree-joint coordinates of all limbs, i.e. $\dot{\bar{%
\mathbold{\vartheta}%
}}_{\left( l\right) },l=1,\ldots ,L-1$ and $\dot{%
\mathbold{\vartheta}%
}_{\left( L\right) }$, in $\dot{\bar{%
\mathbold{\vartheta}%
}}$, and the corresponding inverse kinematics Jacobians, i.e. $\bar{\mathbf{F%
}}_{\left( l\right) },i=1,\ldots ,L-1$ and $\mathbf{F}_{\left( L\right) }$,
of the limbs in $\bar{\mathbf{F}}$, gives rise to the velocity inverse
kinematics solution of the tree-topology mechanism%
\begin{equation}
\dot{\bar{%
\mathbold{\vartheta}%
}}=\bar{\mathbf{F}}\mathbf{V}_{\mathrm{t}}  \label{Feta2}
\end{equation}%
which determines all tree-joint velocities in terms of the task space
velocity satisfying the loop constraints.

The velocity inverse kinematics of the PKM is expressed by means of the
inverse kinematics Jacobian ${\mathbf{J}}_{\mathrm{IK}}$ as

\begin{equation}
\dot{%
\mathbold{\vartheta}%
}_{\mathrm{a}}={\mathbf{J}}_{\mathrm{IK}}\mathbf{V}_{\mathrm{t}}  \label{IK}
\end{equation}%
where vector $%
\mathbold{\vartheta}%
_{\mathrm{a}}$ comprises the coordinates $\vartheta _{i\left( l\right) }$
corresponding to the actuated joints. For non-redundantly actuated PKM, $%
\mathbold{\vartheta}%
_{\mathrm{a}}$ are generalized coordinates. For many PKM with simple limbs
(serial chains), the inverse kinematics Jacobian can be determined easily 
\cite{Merlet2006,BriotKhalilBook} (see Table 6 in~\cite{KUMAR2020102367}).
For PKM with complex limbs, this is more involved \cite%
{MMT2022,MuellerRAS2022}. With the above inverse kinematics solution of the
mechanism (\ref{Feta2}), the inverse kinematics of the PKM is already
available, however. The IK Jacobian consists of the rows of $\bar{\mathbf{F}}
$ corresponding to the actuated joints. This will be exploited when
computing derivatives.

\section{Closed-Form Inverse Dynamics of PKM%
\label{secInvDyn}%
}

To derive the dynamics EOM of the PKM, the EOM of the tree-topology system
introduced above are derived first, and the constraints are imposed using
the above inverse kinematics solution. The EOM of the tree-topology system
split into the EOM of the $L$ individual limbs.

\subsection{Joint Space Formulation of EOM of Individual Limbs}

The EOM of limb $l=1,\ldots ,L$ can be written in the standard form, as for
any tree-topology MBS, as%
\begin{equation}
\bar{\mathbf{M}}_{\left( l\right) }\ddot{\bar{%
\mathbold{\vartheta}%
}}_{\left( l\right) }+\bar{\mathbf{C}}_{\left( l\right) }\dot{\bar{%
\mathbold{\vartheta}%
}}_{\left( l\right) }+\bar{\mathbf{Q}}_{\left( l\right) }^{\mathrm{grav}}=%
\bar{\mathbf{Q}}_{\left( l\right) }  \label{EOMLimb}
\end{equation}%
where $\bar{\mathbf{M}}_{\left( l\right) }(\bar{%
\mathbold{\vartheta}%
}_{\left( l\right) })$ is the generalized mass matrix, $\bar{\mathbf{C}}%
_{\left( l\right) }(\bar{%
\mathbold{\vartheta}%
}_{\left( l\right) },\dot{\bar{%
\mathbold{\vartheta}%
}}_{\left( l\right) })$ the generalized Coriolis/centrifugal matrix, $\bar{%
\mathbf{Q}}_{\left( l\right) }^{\mathrm{grav}}(\bar{%
\mathbold{\vartheta}%
}_{\left( l\right) })$ generalized forces due to gravity, and $\bar{\mathbf{Q%
}}_{\left( l\right) }$ represents all applied forces including actuation
forces. Friction, contact, and other forces are omitted, for simplicity.
Expression (\ref{EOMLimb}) is indeed the EOM of a serial robot. These
equations possess compact closed form expressions that were formalized by
means of the spatial operator algebra \cite%
{Jain1991,JainBook,Featherstone2008}, which is conceptually similar to the
natural orthogonal complement approach \cite{AngelesLee1988,AngelesBook}.
Using matrix Lie group methods, all expressions are intrinsically given in
terms of the screw coordinates and frame transformation matrices \cite%
{LynchPark2017,MUBOScrews2}. These equations can also be evaluated with $%
O\left( n_{l}\right) $ complexity using recursive Lie group algorithms \cite%
{ParkBobrowPloen1995,LynchPark2017} that will be employed for the
higher-order inverse dynamics in Sec. \ref{secFKIK} and \ref{secHighInvdyn}.

\subsection{Task Space Formulation of EOM in Closed Form}

If the PKM is kinematically non-redundant, $\delta =\delta _{\mathrm{p}}$,
the PKM motion is determined by the platform motion. Then the EOM (\ref%
{EOMLimb}) govern the dynamics of a separated limb. The overall task space
formulation of EOM of the PKM is (omitting EE-loads) given in (\ref{EOMTask1}%
) with the $\delta \times \delta $ generalized mass matrix and
Coriolis/centrifugal matrix%
\begin{align}
\mathbf{M}_{\mathrm{t}}(%
\mathbold{\vartheta}%
):=& \sum_{l=1}^{L}\bar{\mathbf{F}}_{\left( l\right) }^{T}\bar{\mathbf{M}}%
_{\left( l\right) }\bar{\mathbf{F}}_{\left( l\right) }  \label{Mt} \\
\mathbf{C}_{\mathrm{t}}(%
\mathbold{\vartheta}%
,\dot{%
\mathbold{\vartheta}%
}):=& \sum_{l=1}^{L}\bar{\mathbf{F}}_{\left( l\right) }^{T}(\bar{\mathbf{C}}%
_{\left( l\right) }\bar{\mathbf{F}}_{\left( l\right) }+\bar{\mathbf{M}}%
_{\left( l\right) }\dot{\bar{\mathbf{F}}}_{\left( l\right) }).  \label{Ct}
\end{align}%
The generalized forces due to gravity are%
\begin{equation}
\mathbf{W}_{\mathrm{t}}^{\mathrm{grav}}(%
\mathbold{\vartheta}%
):=\sum_{l=1}^{L}\bar{\mathbf{F}}_{\left( l\right) }^{T}\bar{\mathbf{Q}}%
_{\left( l\right) }^{\mathrm{grav}}.  \label{Wt}
\end{equation}

\subsection{Inverse Dynamics Formulation}

The inverse dynamics problem is to compute the actuation forces $\mathbf{u}$
for given motion of the PKM and applied wrenches. Instead of using the
closed form EOM (\ref{EOMTask1}), the inverse dynamics solution is expressed
as%
\begin{equation}
\mathbf{u}=\mathbf{J}_{\mathrm{IK}}^{-T}\sum_{l=1}^{L}\bar{\mathbf{F}}%
_{\left( l\right) }^{T}\bar{\mathbf{Q}}_{\left( l\right) }(%
\mathbold{\vartheta}%
_{\left( l\right) },\dot{%
\mathbold{\vartheta}%
}_{\left( l\right) },\ddot{%
\mathbold{\vartheta}%
}_{\left( l\right) })  \label{InvDyn1}
\end{equation}%
using the inverse kinematics solution (\ref{Ftheta1}) and (\ref{Ftheta2}),
with the EOM (\ref{EOMLimb}) of the limbs. The advantage of this form is
that it allows for separate evaluation of the dynamics EOM of the limbs, of
their inverse kinematics solution, and of the inverse kinematics Jacobian $%
\mathbf{J}_{\mathrm{IK}}$, respectively the forward kinematics Jacobian $%
\mathbf{J}_{\mathrm{IK}}^{-1}$. This holds true for their time derivatives,
which allows for an efficient evaluation of the higher-order inverse
dynamics as described in the next section. Note that all terms in (\ref%
{EOMLimb}), and (\ref{InvDyn1}), depend on $%
\mathbold{\vartheta}%
$ and its derivatives.

\section{Fourth-Order Forward/Inverse Kinematics%
\label{secFKIK}%
}

Time derivatives of the inverse dynamics solution (\ref{InvDyn1})
necessitate derivatives of the EOM (\ref{EOMLimb}) as well as of the inverse
kinematics solution (\ref{Ftheta1}) of the limbs. When solving the ordinary
inverse dynamics problem of serial robots, the EOM of the form (\ref{EOMLimb}%
) are evaluated using a recursive $O\left( n_{l}\right) $ algorithm. Any
such inverse dynamics algorithm, which takes joint variables $%
\mathbold{\vartheta}%
$, velocities $\dot{%
\mathbold{\vartheta}%
}$, and accelerations $\ddot{%
\mathbold{\vartheta}%
}$ as inputs, consists of a forward kinematics loop and an inverse dynamics
loop. In the forward kinematics run, the configurations, velocities, and
accelerations of all bodies are computed. Solving this collectively for all
limbs is called the \emph{forward kinematics of the mechanism}. The task
space formulation (\ref{EOMLimb}) additionally involves solving the \emph{%
inverse kinematics problem of the mechanism}, i.e. computing $%
\mathbold{\vartheta}%
,\dot{%
\mathbold{\vartheta}%
},\ddot{%
\mathbold{\vartheta}%
}$ from given platform motion, which means solving the inverse kinematics of
all limbs. The second-order inverse dynamics additionally involves computing
the third- and fourth-order inverse and forward kinematics solution. To this
end, a combined fourth-order forward/inverse kinematics $O\left(
N_{l}\right) $ algorithm is introduced. Computation of derivatives of the
inverse kinematics Jacobian in (\ref{InvDyn1}) is discussed in Sec. \ref%
{secIKder}.

\subsection{Derivatives of the inverse kinematics solution of limbs}

Taking derivatives of (\ref{VpD}), and solving for the highest derivative of 
$%
\mathbold{\vartheta}%
_{\left( l\right) }$, yields the $\nu $-th time derivative of the inverse
kinematics solution (\ref{Ftheta1}) of limb $l$ 
\begin{align}
\frac{d^{\nu }}{dt^{\nu }}%
\mathbold{\vartheta}%
_{\left( l\right) }& =\mathbf{F}_{\left( l\right) }\frac{d^{\nu -1}}{dt^{\nu
-1}}\mathbf{V}_{\mathrm{t}}-\mathbf{J}_{\mathrm{t}\left( l\right) }^{-1}%
\mathbf{P}_{\mathrm{t}\left( l\right) }\mathbf{c}_{\left( l\right) }^{\nu }
\\
\mathrm{with\ \ \ }\mathbf{c}_{\left( l\right) }^{\nu }& :=\sum_{k=1}^{\nu
-1}\tbinom{\nu -1}{k}\frac{d^{k}}{dt^{k}}\mathbf{J}_{\mathrm{p}\left(
l\right) }\frac{d^{\nu -k}}{dt^{\nu -k}}%
\mathbold{\vartheta}%
_{\left( l\right) }.
\end{align}%
For derivatives up to order $\nu =4$, these are%
\begin{align}
\mathbf{c}_{\left( l\right) }^{2}& :=\dot{\mathbf{J}}_{\mathrm{p}\left(
l\right) }\dot{%
\mathbold{\vartheta}%
}_{\left( l\right) },\ \mathbf{c}_{\left( l\right) }^{3}:=\ddot{\mathbf{J}}_{%
\mathrm{p}\left( l\right) }\dot{%
\mathbold{\vartheta}%
}_{\left( l\right) }+2\dot{\mathbf{J}}_{\mathrm{p}\left( l\right) }\ddot{%
\mathbold{\vartheta}%
}_{\left( l\right) }  \label{c3} \\
\mathbf{c}_{\left( l\right) }^{4}& :=\dddot{\mathbf{J}}_{\mathrm{p}\left(
l\right) }\dot{%
\mathbold{\vartheta}%
}_{\left( l\right) }+2\ddot{\mathbf{J}}_{\mathrm{p}\left( l\right) }\ddot{%
\mathbold{\vartheta}%
}_{\left( l\right) }+3\dot{\mathbf{J}}_{\mathrm{p}\left( l\right) }\dddot{%
\mathbold{\vartheta}%
}_{\left( l\right) }.  \label{c4}
\end{align}%
Computing inverse kinematics derivatives boils down to the inversion of the
task space Jacobian and computing derivatives of the forward kinematics
Jacobians of the limbs. They admit the closed form expressions (\ref{Jpd}-%
\ref{Jpddd}), see appendix.

\subsection{Recursive forward/inverse kinematics algorithm}

The second-order inverse dynamics involves solving the 4th-order inverse
kinematics of the mechanism (computing $\dot{%
\mathbold{\vartheta}%
}_{\left( l\right) },\ldots ,\ddot{\ddot{%
\mathbold{\vartheta}%
}}_{\left( l\right) }$ from $\mathbf{V}_{\mathrm{t}},\ldots ,\dddot{\mathbf{V%
}}_{\mathrm{t}}$) and the forward kinematics of the mechanism (computing all
body twists $\mathbf{V}_{i\left( l\right) },\ldots ,\dddot{\mathbf{V}}%
_{i\left( l\right) }$ from $\dot{%
\mathbold{\vartheta}%
}_{\left( l\right) },\ldots ,\ddot{\ddot{%
\mathbold{\vartheta}%
}}_{\left( l\right) }$). Both are computed together in a single run.
Recursive algorithms for computing the fourth-order forward kinematics of
serial chains were reported in \cite{BuondonnaDeLuca2016,Guarino2009} using
classical vector formulations. In the following, the algorithm introduced in 
\cite{ICRA2017} that uses the Lie group formulation from \cite{MUBOScrews1}
is complemented with the inverse kinematics. An introduction to the Lie
group formulation of serial chains in general can be found in \cite%
{LynchPark2017}. A brief summary of the notation is given in the appendix.
The homogenous transformation $\mathbf{C}_{i\left( l\right) }$ (not to be
confused with $\mathbf{C}_{\mathrm{t}}$ in (\ref{Ct})) represents the
absolute configuration of body $i$ of limb $l$ relative to the inertial
frame, $\mathbf{C}_{i,i-1}$ is the relative configuration of body $i$ and $%
i-1$, and $\mathbf{C}_{\mathrm{p}}$ denotes the absolute platform
configuration relative to the inertial frame. 

The algorithm consists of four subsequent loops. Input to the first loop is
the current PKM state $(%
\mathbold{\vartheta}%
,\mathbf{V}_{\mathrm{t}})$. Solving the geometric inverse kinematics
problem, computing $%
\mathbold{\vartheta}%
$ from given platform motion configuration $\mathbf{C}_{\mathrm{p}}$, is not
the subject here, and $%
\mathbold{\vartheta}%
$ is assumed to be given. Denote with $\mathbf{g}$ the vector of
gravitational acceleration expressed in inertial frame, then the
corresponding acceleration screw of the ground is $\dot{\mathbf{V}}_{0\left(
l\right) }=\left( \mathbf{0},\mathbf{g}\right) ^{T}$. The algorithm is
derived by splitting the forward kinematics run presented in \cite{ICRA2017}%
, necessary as the inverse kinematics must be solved for each order first.
Matrix $\mathbf{Ad}_{\mathbf{C}_{i,i-1}}$ transforms a twist from body $i-1$
to body $i$, and $\mathbf{ad}_{\mathbf{X}}\mathbf{Y}$ yields the Lie bracket
of $\mathbf{X},\mathbf{Y}\in {\mathbb{R}}^{6}$, also called 'spatial cross
product' 
\'{}%
\cite{Featherstone2008,JainBook}. Notice that $n_{L}=N_{L}$.

\parindent0pt%
\vspace{1ex}%

\underline{1st-order Inverse and 0th-order Forward Kinematics}%
\vspace{-0.5ex}%

\begin{itemize}
\item Input: $%
\mathbold{\vartheta}%
,\mathbf{V}_{\mathrm{t}}$

\item FOR $l=1,\ldots ,L$ DO (possible in parallel)

\begin{itemize}
\item[ ] FOR $i=1,\ldots ,N_{l}$ (omitting subscript $\left( l\right) $)

\begin{itemize}
\item[ ] $\mathbf{C}_{i}=\ \mathbf{C}_{i-1}\mathbf{B}_{i}\exp ({\mathbf{X}}%
_{i}\vartheta _{i})$
\end{itemize}

compute $\mathbf{J}_{\mathrm{p}\left( l\right) }$ with (\ref{Jp})

$\mathbf{F}_{\left( l\right) }=\mathbf{J}_{\mathrm{t}\left( l\right) }^{-1}%
\mathbf{D}_{\mathrm{t}\left( l\right) }$

$\dot{%
\mathbold{\vartheta}%
}_{\left( l\right) }=\mathbf{F}_{\left( l\right) }\mathbf{V}_{\mathrm{t}}%
\hfill%
(\ast )$%
\vspace{-1ex}%
\end{itemize}

\item Output: $\mathbf{J}_{\mathrm{p}\left( l\right) },\mathbf{F}_{\left(
l\right) },\dot{%
\mathbold{\vartheta}%
},\mathbf{C}_{i\left( l\right) }$
\end{itemize}

\parindent0pt%

\underline{2nd-order Inverse and 1st-order Forward Kinematics}%
\vspace{-0.5ex}%

\begin{itemize}
\item Input: $%
\mathbold{\vartheta}%
,\dot{%
\mathbold{\vartheta}%
},\dot{\mathbf{V}}_{\mathrm{t}},\mathbf{J}_{\mathrm{p}\left( l\right) },%
\mathbf{F}_{\left( l\right) },\mathbf{C}_{i\left( l\right) }$

\item FOR $l=1,\ldots ,L$ DO (possible in parallel)

\begin{itemize}
\item[ ] FOR $i=1,\ldots ,N_{l}$ (omitting subscript $\left( l\right) $)

$\ \ \ \mathbf{V}_{i}=\ \mathbf{Ad}_{\mathbf{C}_{i,i-1}}\mathbf{V}_{i-1}+{%
\mathbf{X}}_{i}\dot{\vartheta}_{i}$

\ \ \ compute $\dot{\mathbf{J}}_{\mathrm{p}\left( l\right) ,i}$ with (\ref%
{Jpd})

compute $\mathbf{c}_{\left( l\right) }^{2}$ with (\ref{c3})

$\ddot{%
\mathbold{\vartheta}%
}_{\left( l\right) }=\mathbf{F}_{\left( l\right) }%
\big%
(\dot{\mathbf{V}}_{\mathrm{t}}-\mathbf{J}_{\mathrm{t}\left( l\right) }^{-1}%
\mathbf{P}_{\mathrm{t}\left( l\right) }\mathbf{c}_{\left( l\right) }^{2}%
\big%
)%
\hfill%
(\ast )$
\end{itemize}

\item Output: $\ddot{%
\mathbold{\vartheta}%
},\mathbf{V}_{i}$

\parindent0pt%
\vspace{1ex}%
\end{itemize}

\underline{3rd-order Inverse and 2nd-order Forward Kinematics}%
\vspace{-0.5ex}%

\begin{itemize}
\item Input: $%
\mathbold{\vartheta}%
,\dot{%
\mathbold{\vartheta}%
},\ddot{%
\mathbold{\vartheta}%
},\ddot{\mathbf{V}}_{\mathrm{t}},\mathbf{J}_{\mathrm{p}\left( l\right) },%
\mathbf{F}_{\left( l\right) },\mathbf{C}_{i\left( l\right) },\mathbf{V}%
_{i\left( l\right) }$

\item FOR $l=1,\ldots ,L$ DO (possible in parallel)

\begin{itemize}
\item[ ] FOR $i=1,\ldots ,N_{l}$ (omitting subscript $\left( l\right) $)

$\ \ \ \dot{\mathbf{V}}_{i}=\ \mathbf{Ad}_{\mathbf{C}_{i,i-1}}\dot{\mathbf{V}%
}_{i-1}-\dot{\vartheta}_{i}\mathbf{ad}_{{\mathbf{X}}_{i}}\mathbf{V}_{i}+{%
\mathbf{X}}_{i}\ddot{\vartheta}_{i}$

\ \ \ compute $\ddot{\mathbf{J}}_{\mathrm{p}\left( l\right) ,i}$ with (\ref%
{Jpdd})

compute $\mathbf{c}_{\left( l\right) }^{3}$ with (\ref{c3})

$\dddot{%
\mathbold{\vartheta}%
}=\mathbf{F}_{\left( l\right) }%
\big%
(\ddot{\mathbf{V}}_{\mathrm{t}}-\mathbf{J}_{\mathrm{t}\left( l\right) }^{-1}%
\mathbf{P}_{\mathrm{t}\left( l\right) }\mathbf{c}_{\left( l\right) }^{3}%
\big%
)%
\hfill%
(\ast )$
\end{itemize}

\item Output: $\dddot{%
\mathbold{\vartheta}%
},\dot{\mathbf{V}}_{i\left( l\right) }$

\parindent0pt%
\vspace{1ex}%
\end{itemize}

\underline{4th-order Inverse and 3rd-order Forward Kinematics}%
\vspace{-0.5ex}%

\begin{itemize}
\item Input: $%
\mathbold{\vartheta}%
,\dot{%
\mathbold{\vartheta}%
},\ddot{%
\mathbold{\vartheta}%
},\dddot{%
\mathbold{\vartheta}%
},\dddot{\mathbf{V}}_{\mathrm{t}},\mathbf{J}_{\mathrm{p}\left( l\right) },%
\mathbf{F}_{\left( l\right) },\mathbf{C}_{i\left( l\right) },\mathbf{V}%
_{i\left( l\right) },\dot{\mathbf{V}}_{i\left( l\right) }$

\item FOR $l=1,\ldots ,L$ DO (possible in parallel)

\begin{itemize}
\item[ ] FOR $i=1,\ldots ,N_{l}$ (omitting subscript $\left( l\right) $)

$\ \ \ \ddot{\mathbf{V}}_{i}=\ \mathbf{Ad}_{\mathbf{C}_{i,i-1}}\ddot{\mathbf{%
V}}_{i-1}+{\mathbf{X}}_{i}\dddot{\vartheta}_{i}$

$\ \ \ \ \ \ \ \ \ \ \ -\mathbf{ad}_{{\mathbf{X}}_{i}}(\ddot{\vartheta}_{i}%
\mathbf{V}_{i}+2\dot{\vartheta}_{i}\dot{\mathbf{V}}_{i})-\dot{\vartheta}%
_{i}^{2}\mathbf{ad}_{{\mathbf{X}}_{i}}^{2}\mathbf{V}_{i}$

\ \ \ compute $\dddot{\mathbf{J}}_{\mathrm{p}\left( l\right) ,i}$ with (\ref%
{Jpddd})

compute $\mathbf{c}_{\left( l\right) }^{4}$ with (\ref{c4})

$\ddot{\ddot{%
\mathbold{\vartheta}%
}}=\mathbf{F}_{\left( l\right) }%
\big%
(\dddot{\mathbf{V}}_{\mathrm{t}}-\mathbf{J}_{\mathrm{t}\left( l\right) }^{-1}%
\mathbf{P}_{\mathrm{t}\left( l\right) }\mathbf{c}_{\left( l\right) }^{4}%
\big%
)%
\hfill%
(\ast )$
\end{itemize}

\item Output: $\ddot{\ddot{%
\mathbold{\vartheta}%
}},\ddot{\mathbf{V}}_{i\left( l\right) }$

\parindent0pt%
\vspace{1ex}%
\end{itemize}

\underline{4th-order Forward Kinematics}%
\vspace{-0.5ex}%

\begin{itemize}
\item Input: $%
\mathbold{\vartheta}%
,\dot{%
\mathbold{\vartheta}%
},\ddot{%
\mathbold{\vartheta}%
},\dddot{%
\mathbold{\vartheta}%
},\ddot{\ddot{\mbox{$\vartheta$}}}_{i},\mathbf{J}_{\mathrm{p}\left( l\right)
},\mathbf{F}_{\left( l\right) },\mathbf{C}_{i\left( l\right) },\mathbf{V}%
_{i\left( l\right) },\dot{\mathbf{V}}_{i\left( l\right) },\ddot{\mathbf{V}}%
_{i\left( l\right) }$

\item FOR $l=1,\ldots ,L$ DO (possible in parallel)

\begin{itemize}
\item[ ] FOR $i=1,\ldots ,N_{l}$%
\vspace{-1ex}%
\begin{align*}
\dddot{\mathbf{V}}_{i}=& \ \mathbf{Ad}_{\mathbf{C}_{i,i-1}}\dddot{\mathbf{V}}%
_{i-1}+{\mathbf{X}}_{i}\ddot{\ddot{\mbox{$\vartheta$}}}_{i} \\
& -\mathbf{ad}_{{\mathbf{X}}_{i}}(\dddot{\vartheta}_{i}\mathbf{V}_{i}+3\ddot{%
\vartheta}_{i}\dot{\mathbf{V}}_{i}+3\dot{\vartheta}_{i}\ddot{\mathbf{V}}_{i})
\\
& -3\dot{\vartheta}_{i}\mathbf{ad}_{{\mathbf{X}}_{i}}^{2}(\ddot{\vartheta}%
_{i}\mathbf{V}_{i}+\dot{\vartheta}_{i}\dot{\mathbf{V}}_{i})-\dot{\vartheta}%
_{i}^{3}\mathbf{ad}_{{\mathbf{X}}_{i}}^{3}\mathbf{V}_{i}%
\vspace{-1ex}%
\end{align*}
\end{itemize}

\item Output: $\dddot{\mathbf{V}}_{i\left( l\right) }$
\end{itemize}

\subsection{Computational complexity}

In contrast to the forward kinematics of serial robots, prior to the $\nu $%
th-order forward kinematics, the $\nu $th-order inverse kinematic problem is
solved (for which the ($\nu $-1)st time derivative of $\mathbf{J}_{\mathrm{t}%
\left( l\right) }$ is computed) separately for each order. The $\nu $%
th-order kinematics run ($D^{\left( \nu \right) }%
\mathbold{\vartheta}%
\mapsto D^{\left( \nu -1\right) }\mathbf{V}_{i\left( l\right) }$) for limb $l
$ has complexity $O\left( N_{l}\right) $. Each run involves an expression $%
\mathbf{a}=\mathbf{F}_{\left( l\right) }\mathbf{b}$ with $\mathbf{F}_{\left(
l\right) }:=\mathbf{J}_{\mathrm{t}\left( l\right) }^{-1}\mathbf{D}_{\mathrm{t%
}\left( l\right) }$, indicted with (*), solving the $\nu $th-order inverse
kinematics problem ($D^{\left( \nu -1\right) }\mathbf{V}_{\mathrm{t}}\mapsto
D^{\left( \nu \right) }%
\mathbold{\vartheta}%
_{\left( l\right) }$). Inversion of the $\delta _{\mathrm{p}}\times \delta _{%
\mathrm{p}}$ task space Jacobians is avoided by solving $\mathbf{J}_{\mathrm{%
t}\left( l\right) }\mathbf{b}=\mathbf{D}_{\mathrm{t}\left( l\right) }\mathbf{%
a}$ for $\mathbf{b}$. LU-decompositions of $\mathbf{J}_{\mathrm{t}\left(
l\right) }$ are computed before the kinematics runs, and reused for solving
the four equations (*). The total number of operations is $L\left( \frac{2}{3%
}\delta _{\mathrm{p}}^{3}+8\delta _{\mathrm{p}}^{2}\right) $. The overall
complexity of the kinematics run is $O\left( L\cdot N_{l}\right) $, except
for solving (*).

\section{Second-Order Inverse Dynamics%
\label{secHighInvdyn}%
}

\subsection{Derivatives of the inverse dynamics solution}

The first and second time derivative of the inverse dynamics solution (\ref%
{InvDyn1}) are readily found as (omitting EE loads)%
\begin{equation}
\dot{\mathbf{u}}=\mathbf{J}_{\mathrm{IK}}^{-T}%
\Big%
(\sum_{l=1}^{L}%
\big%
(\mathbf{F}_{\left( l\right) }^{T}\dot{\bar{\mathbf{Q}}}_{\left( l\right) }+%
\dot{\mathbf{F}}_{\left( l\right) }^{T}\bar{\mathbf{Q}}_{\left( l\right) }%
\big%
)-\dot{\mathbf{J}}_{\mathrm{IK}}^{T}\mathbf{u}%
\Big%
)  \label{ud}
\end{equation}%
\begin{align}
\ddot{\mathbf{u}}& =\mathbf{J}_{\mathrm{IK}}^{-T}%
\Big%
(\sum_{l=1}^{L}%
\big%
(\mathbf{F}_{\left( l\right) }^{T}\ddot{\bar{\mathbf{Q}}}_{\left( l\right)
}+2\dot{\mathbf{F}}_{\left( l\right) }^{T}\dot{\bar{\mathbf{Q}}}_{\left(
l\right) }+\ddot{\mathbf{F}}_{\left( l\right) }^{T}\bar{\mathbf{Q}}_{\left(
l\right) }%
\big%
)  \notag \\
& \ \ \ \ \ \ \ \ \ \ \ -\ddot{\mathbf{J}}_{\mathrm{IK}}^{T}\mathbf{u}-2\dot{%
\mathbf{J}}_{\mathrm{IK}}^{T}\dot{\mathbf{u}}\Big)  \label{udd}
\end{align}%
with $\dot{\mathbf{F}}_{\left( l\right) }=-\mathbf{F}_{\left( l\right) }\dot{%
\mathbf{J}}_{\mathrm{t}\left( l\right) }\mathbf{F}_{\left( l\right) }$ and $%
\ddot{\mathbf{F}}_{\left( l\right) }=2\dot{\mathbf{F}}_{\left( l\right) }%
\dot{\mathbf{J}}_{\mathrm{t}\left( l\right) }\dot{\mathbf{F}}_{\left(
l\right) }-\mathbf{F}_{\left( l\right) }\ddot{\mathbf{J}}_{\mathrm{t}\left(
l\right) }\mathbf{F}_{\left( l\right) }$. Noting that $\mathbf{J}_{\mathrm{t}%
\left( l\right) }:=\mathbf{P}_{\mathrm{t}\left( l\right) }\mathbf{J}_{%
\mathrm{p}\left( l\right) }$, the latter are available with derivatives of $%
\mathbf{J}_{\mathrm{p}\left( l\right) }$ in (\ref{Jpd})-(\ref{Jpddd}) in
appendix.

\subsection{Derivatives of manipulator inverse kinematics Jacobian%
\label{secIKder}%
}

The manipulator inverse kinematics Jacobian $\mathbf{J}_{\mathrm{IK}}$
consists of the $\delta $ rows of the inverse kinematics Jacobians of the $l$
limbs. That is, each row of $\mathbf{F}_{\left( l\right) }$ that corresponds
to an actuated joint delivers one row of $\mathbf{J}_{\mathrm{IK}}^{-1}$. If
each limb comprises one actuator, then $L=\delta $, and each $\mathbf{F}%
_{\left( l\right) }$ contributes one row. Thus the derivatives of $\mathbf{J}%
_{\mathrm{IK}}$ are already available with the derivatives of $\mathbf{F}%
_{\left( l\right) }$ above.

\subsection{Recursive 2nd-Order $O\left( n_{l}\right) $ Inverse Dynamics
Algorithm}

Evaluating (\ref{udd}) involves computing derivatives of $\bar{\mathbf{Q}}%
_{\left( l\right) }$, i.e. of the second-oder inverse dynamics solution of a
serial kinematic chain. The advantage of the proposed higher-order inverse
dynamics method for PKM is that this evaluation is separated from the
kinematics modeling. Thus any computation scheme or software that delivers
the inverse dynamics solution derivatives can be used (which may have any
level of modeling detail including flexibilities etc). Various recursive
algorithms have been proposed to evaluate the second time derivatives of the
EOM (\ref{EOMLimb}) of the limbs, e.g. \cite{BuondonnaDeLuca2016,Guarino2009}%
. Lie group formulations were proposed in \cite{ICRA2017,RAL2020}, which are
coordinate-invariant and compact, and thus easy to implement. In the
following, the inverse dynamics run of the algorithm in \cite{ICRA2017} is
adopted. The forward kinematics run is accomplished by the algorithm in Sec. %
\ref{secFKIK}. Denoting with $\mathbf{M}_{i}$ the $6\times 6$ mass matrix of
body $i$ expressed in the (arbitrary) body-fixed frame, and the interbody
wrenches with $\bar{\mathbf{W}}_{i}$ (merely algorithmic variables), the
inverse dynamics run is:%
\vspace{1ex}%

\underline{Inverse Dynamics}%
\vspace{-0.5ex}%

\begin{itemize}
\item Input: $\mathbf{C}_{i\left( l\right) },\mathbf{V}_{i\left( l\right) },%
\dot{\mathbf{V}}_{i\left( l\right) },\ddot{\mathbf{V}}_{i\left( l\right) },%
\dddot{\mathbf{V}}_{i\left( l\right) }$

\item FOR $l=1,\ldots ,L$ DO (possible in parallel)

\begin{itemize}
\item[ ] FOR $i=N_{l}-1,\ldots ,1$ (omitting subscript $\left( l\right) $)%
\vspace{-1ex}%
\begin{eqnarray*}
\hspace{1ex}%
\bar{\mathbf{W}}_{i}%
\hspace{-1.6ex}
&=&%
\hspace{-1.6ex}%
\mathbf{Ad}_{\mathbf{C}_{i+1,i}}^{T}\bar{\mathbf{W}}_{i+1}+\mathbf{M}_{i}%
\dot{\mathbf{V}}_{i}-\mathbf{ad}_{\mathbf{V}_{i}}^{T}\mathbf{M}_{i}\mathbf{V}%
_{i} \\
\dot{\bar{\mathbf{W}}}_{i}%
\hspace{-1.6ex}
&=&%
\hspace{-1.6ex}%
\mathbf{Ad}_{\mathbf{C}_{i+1,i}}^{T}(\dot{\bar{\mathbf{W}}}_{i+1}-\dot{%
\vartheta}_{i+1}\mathbf{ad}_{{\mathbf{X}}_{i+1}}^{T}\bar{\mathbf{W}}_{i+1})
\\
&&%
\hspace{-1.6ex}%
+\,\mathbf{M}_{i}\ddot{\mathbf{V}}_{i}-\mathbf{ad}_{\mathbf{V}_{i}}^{T}%
\mathbf{M}_{i}\dot{\mathbf{V}}_{i}-\mathbf{ad}_{\dot{\mathbf{V}}_{i}}^{T}%
\mathbf{M}_{i}\mathbf{V}_{i} \\
\ddot{\bar{\mathbf{W}}}_{i}%
\hspace{-1.6ex}
&=&%
\hspace{-1.6ex}%
\mathbf{Ad}_{\mathbf{C}_{i+1,i}}^{T}%
\Big%
(\ddot{\bar{\mathbf{W}}}_{i+1}-2\dot{\vartheta}_{i+1}\mathbf{ad}_{{\mathbf{X}%
}_{i+1}}^{T}\dot{\bar{\mathbf{W}}}_{i+1} \\
&&%
\hspace{-1.6ex}%
+\,(\dot{\vartheta}_{i+1}^{2}\mathbf{ad}_{{\mathbf{X}}_{i+1}}^{2T}-\ddot{%
\vartheta}_{i+1}\mathbf{ad}_{{\mathbf{X}}_{i+1}}^{T})\bar{\mathbf{W}}_{i+1}%
\Big%
) \\
&&%
\hspace{-1.6ex}%
+\,\mathbf{M}_{i}\dddot{\mathbf{V}}_{i}-\mathbf{ad}_{\mathbf{V}_{i}}^{T}%
\mathbf{M}_{i}\ddot{\mathbf{V}}_{i}-\mathbf{ad}_{\ddot{\mathbf{V}}_{i}}^{T}%
\mathbf{M}_{i}\mathbf{V}_{i} \\
&&%
\hspace{-1.6ex}%
-\,2\mathbf{ad}_{\dot{\mathbf{V}}_{i}}^{T}\mathbf{M}_{i}\dot{\mathbf{V}}_{i}
\\
\bar{Q}_{i}%
\hspace{-1.6ex}
&=&%
\hspace{-1.6ex}%
{\mathbf{X}}_{i}^{T}\bar{\mathbf{W}}_{i},\ \ \dot{\bar{Q}}_{i}={\mathbf{X}}%
_{i}^{T}\dot{\bar{\mathbf{W}}}_{i},\ \ddot{\bar{Q}}_{i}={\mathbf{X}}_{i}^{T}%
\ddot{\bar{\mathbf{W}}}_{i}
\end{eqnarray*}
\end{itemize}

\item Output: $\bar{\mathbf{Q}}_{\left( l\right) },\dot{\bar{\mathbf{Q}}}%
_{\left( l\right) },\ddot{\bar{\mathbf{Q}}}_{\left( l\right) }$%
\vspace{2ex}%
\end{itemize}

\subsection{Computational Aspects}

The computational effort for evaluating the inverse dynamics of limb $l$ is
of order $O\left( n_{l}\right) $, with $n_{L}=N_{L}$. The inverse dynamics
run is thus of order $O\left( L\cdot n_{l}\right) $. The special topology of
PKM with structurally identical limbs can be exploited for an efficient
modular modeling, where the kinematic and dynamic EOM of a prototypical limb
are reused for all limbs \cite{MuellerAMR2020}. The recursive algorithm, it
can be implemented once for a prototypical limb, and $L$ instances are used
during the computation. A simple example is shown in \cite{EUCOMES2020}.

\section{Examples and Simulation Results%
\label{secExamples}%
}

Implementation results are presented when the proposed algorithm is applied
to a 6 DOF GSP and 2 DOF planar PKM. The algorithm was implemented in MATLAB.%
%
%
%
%
%
%
%

\subsection{Gough-Stewart Platform}

\begin{figure}[b]
\centering%
\begin{subfigure}[b]{0.55\linewidth}
         \centering
         \includegraphics[width=\textwidth]{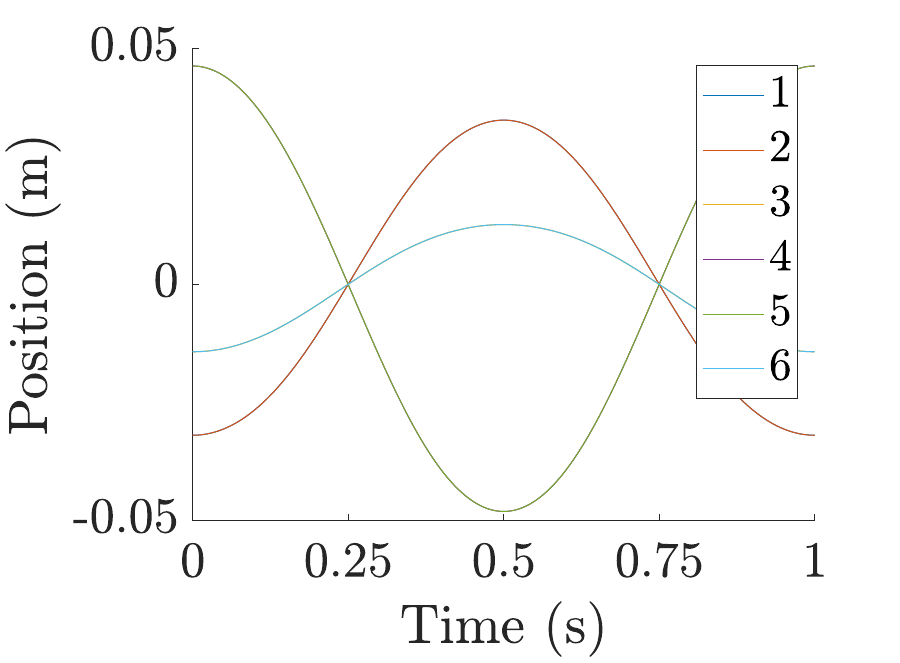}
         \caption{Inv. kinematics solution $\mathbold{\vartheta}_{\rm a}(t)$}
         \label{fig_act_pos}
     \end{subfigure}%
\begin{subfigure}[b]{0.45\linewidth}
         \centering
         \includegraphics[width=\textwidth]{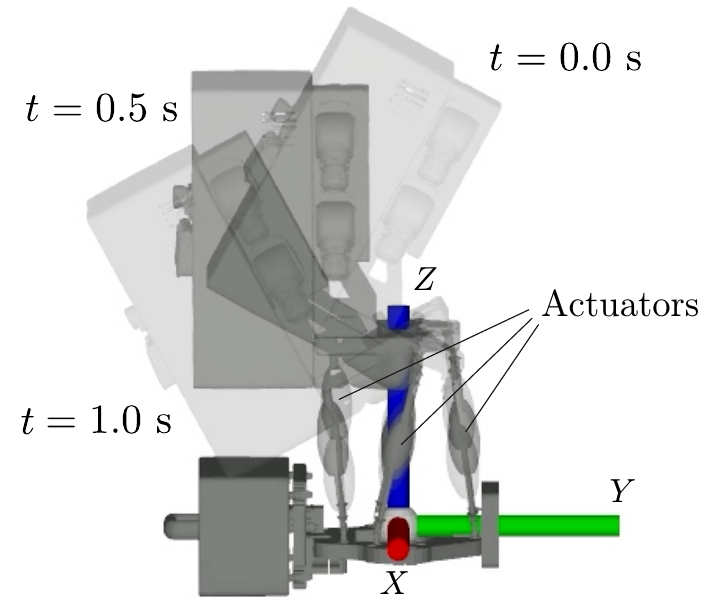}
         \caption{Motion visualization}
         \label{fig_anim}
     \end{subfigure}
\caption{Inverse kinematics solution and animation of Gough-Stewart module
of the Recupera-Reha exoskeleton.}
\label{fig_stewart_animation}
\end{figure}
\begin{figure}[tbh]
\centering%
\begin{subfigure}[b]{0.49\linewidth}
         \centering
         \includegraphics[width=\textwidth]{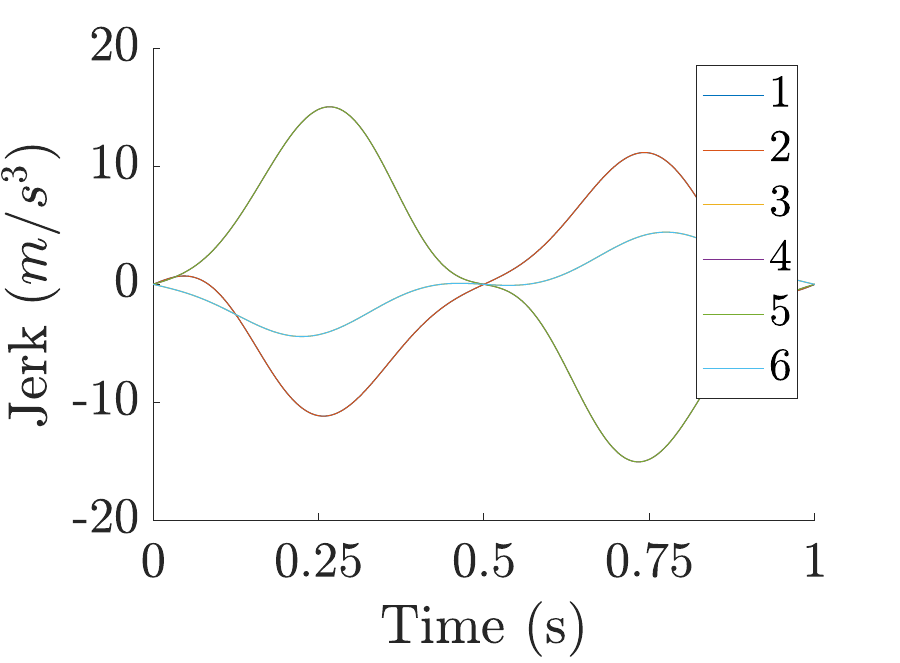}
         \caption{Jerk $\dddot{\mathbold{\vartheta}}_{\rm a}$}
         \label{fig_act_jrk}
     \end{subfigure}
\begin{subfigure}[b]{0.49\linewidth}
         \centering
         \includegraphics[width=\textwidth]{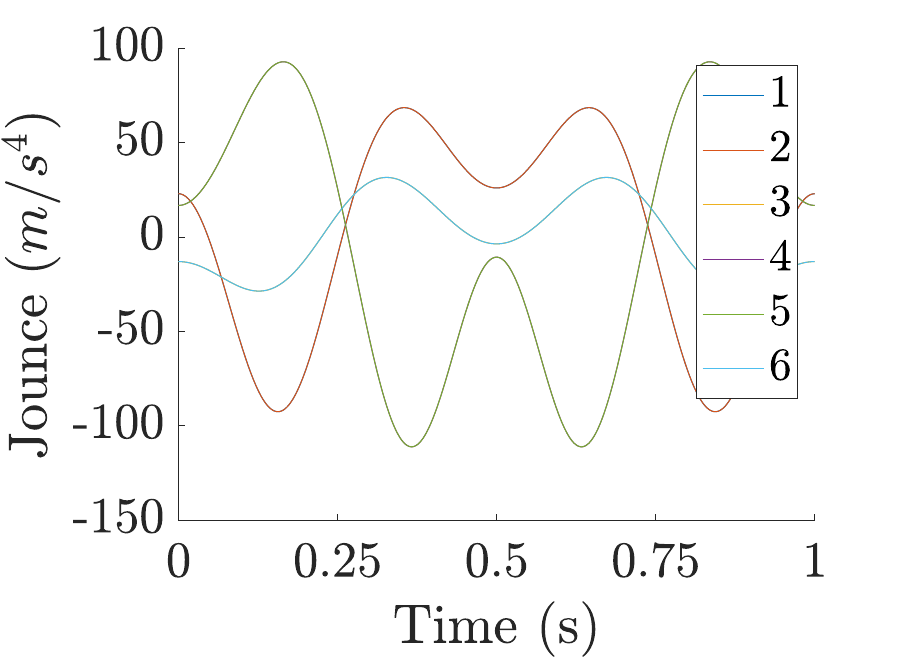}
         \caption{Jounce $\ddot{\ddot{\mathbold{\vartheta}}}_{\rm a}$}
         \label{fig_act_jnc}
     \end{subfigure}
\caption{Higher-order inverse kinematics solution.}
\label{fig_act_traj}
\end{figure}
\begin{figure}[tbh]
\centering%
\begin{subfigure}[b]{0.5\linewidth}
     \centering
     \includegraphics[width=\textwidth]{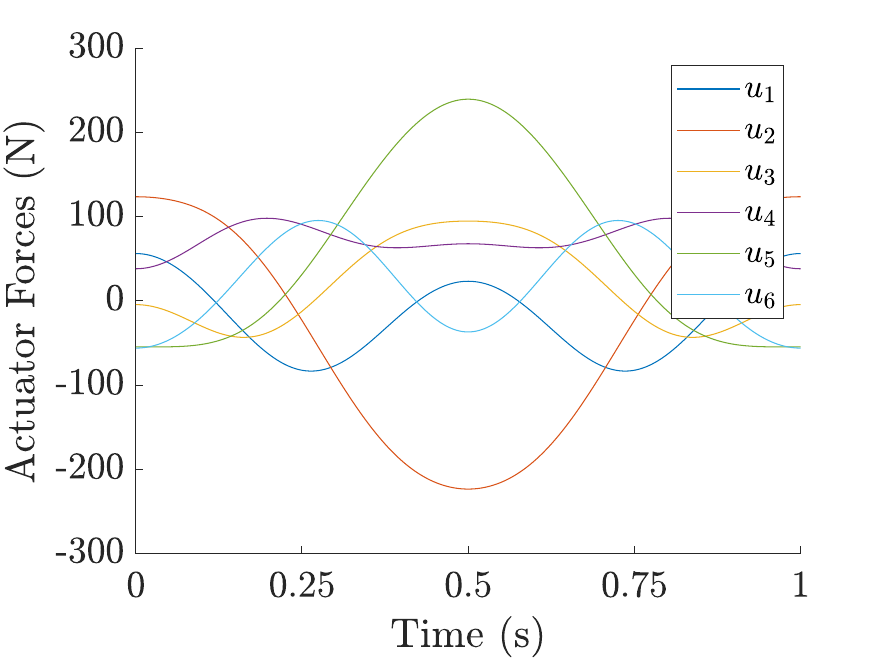}
     \caption{Actuator forces ${\bf u}(t)$}
     \label{fig_act_force}
 \end{subfigure}%
\begin{subfigure}[b]{0.5\linewidth}
         \centering
         \includegraphics[width=\textwidth]{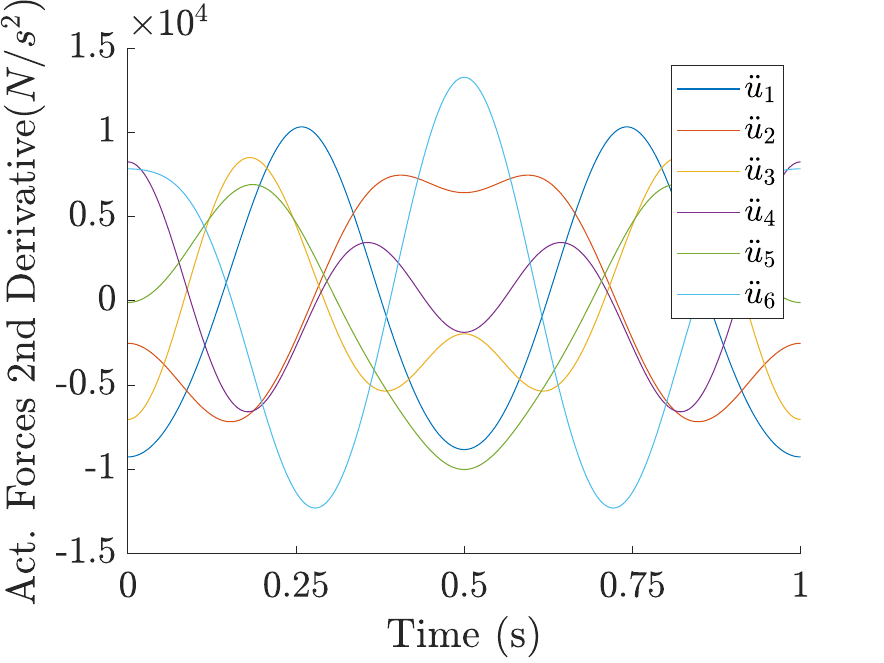}
         \caption{2nd time derivatives $\ddot{\bf u}(t)$}
         \label{fig_act_force_ddot}
\end{subfigure}
\caption{Higher-order inverse dynamics results.}
\label{fig_idyn}
\end{figure}

The proposed method is applied to compute the second-order inverse dynamics
of a 6U\underline{P}S GSP, which is used within the torso of the
Recupera-Reha exoskeleton~\cite{app9040626}. The topology information, joint
screw coordinate vectors ${}\mathbf{X}_{i}$, the relative reference
configurations $\mathbf{B}_{i}$ of all links, and the inertia data $\mathbf{M%
}_{i}$ w.r.t. to the body-fixed frames are extracted from the URDF
description of the PKM. A pure roll motion of the platform parameterized as $%
\theta \left( t\right) =-A\cos (\omega t)+A+\theta _{\text{min}}$ with
magnitude $A=(\theta _{\text{min}}\ -\theta _{\text{max}})/2$ and frequency $%
\omega =\frac{2\pi }{T}$, where $\theta _{\text{min}}=-0.5$\thinspace rad,
and $\theta _{\text{max}}=0.5$\thinspace rad are the minimum and maximum of
the roll angle $\theta $, and $T=1$\thinspace s is the cycle time. The
actuator trajectories ${\mathbold{\vartheta}}_{\mathrm{a}}(t)$ obtained via
the inverse kinematics and the visualization of mechanism motion are shown
in Fig. \ref{fig_act_pos} and \ref{fig_anim}, respectively, and their 3rd
and 4th time derivatives in Fig. The actuator forces $\mathbf{u}(t)$ and
their second time derivatives $\ddot{\mathbf{u}}(t)$ are shown in \ref%
{fig_act_traj}. Fig. \ref{fig_idyn}, respectively. The solutions $\mathbf{u}%
(t)$ and ${\mathbold{\vartheta}}_{\mathrm{a}}\left( t\right) ,\dot{{%
\mathbold{\vartheta}}}_{\mathrm{a}}\left( t\right) ,\ddot{{%
\mathbold{\vartheta}}}_{\mathrm{a}}(t)$ of the inverse kinematics
respectively inverse dynamics are computed with the already existing and
validated model. The higher order inverse kinematics and inverse dynamics
results were verified by numerically differentiating $\mathbf{u}(t)$ and $%
\ddot{{\mathbold{\vartheta}}}(t)$ twice. The $O\left( n\right) $-solution
and the numerical derivatives agree up to machine precision. As an
indication of the computational performance, the total CPU time spent for
10000 evaluations of the MATLAB implementation of the second order inverse
dynamics was measured on a standard laptop computer with Intel Core
i7-4810MQ CPU at 2.8 GHz. Executing 10000 calls, the average CPU time per
call was 0.75\thinspace ms, which is sufficient for real-time control
applications. An order of magnitude reduction is expected from an optimized
C++ implementation. It is to be noted that no parallelization opportunities
were exploited. It will also be interesting to compare the performance with
dedicate MBS codes. When modeled as MBS in terms of relative coordinates,
this example has a spanning tree with 21 DOF subjected to 15 independent
constraints.

\subsection{2-DOF planar 3\protect\underline{R}RR PKM}

\begin{figure}[tbh]
\vspace{-2ex} \centering
\includegraphics[width=0.23\textwidth]{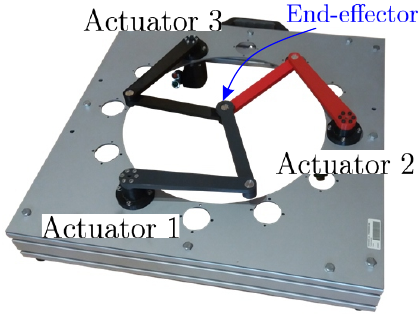}~~ %
\includegraphics[width=0.2\textwidth]{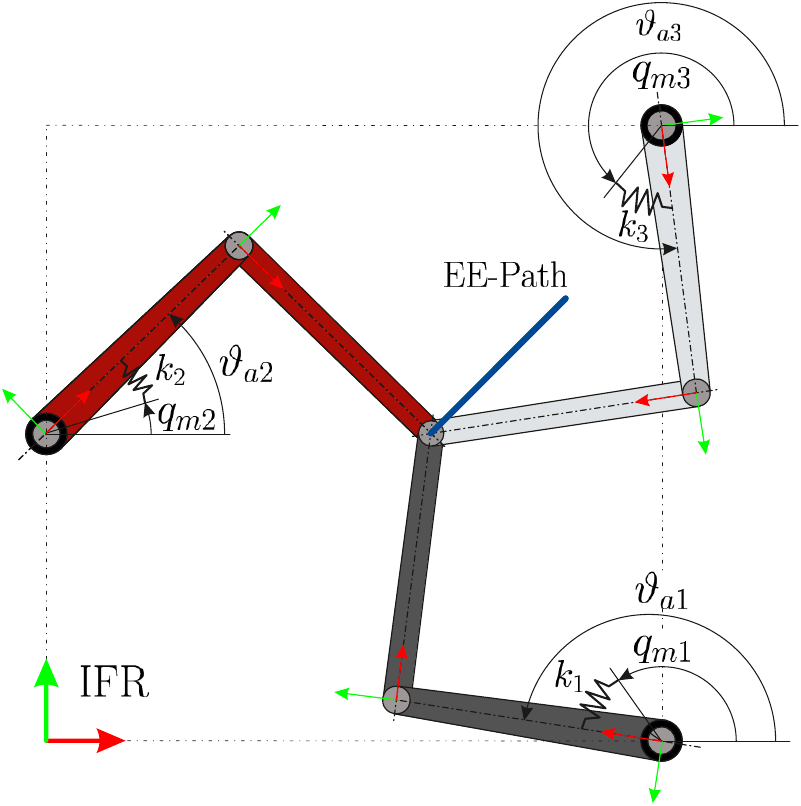}
\caption{Planar 2-DOF 3-\protect\underline{R}RR SEA-PKM. Physical prototype
and schematic drawing. \protect\vspace{-3ex}}
\label{fig_2RRR_PKM}
\end{figure}

The SEA driven 2-DOF planar PKM in Fig. \ref{fig_2RRR_PKM} is considered to
demonstrate the flatness-based control. The position of the EE in the plane
of motion is the output of this PKM, described by its $\left( x,y\right) $%
-coordinates relative to the shown inertial frame (IFR). The PKM is
controlled by three actuators, and would thus be redundantly actuated if the
actuators were kinematically affixed to the arms. Instead, each arm is
mounted on the output shaft of a SEA. The elastic coupling of output shaft
and motor-gear unit is modeled by a torsion spring with stiffness constant $%
k_{i}$ as indicated in Fig. \ref{fig_2RRR_PKM}. Geometric and inertia
parameter are deduced from the CAD model, and motor specifications. The
desired trajectory represents a pick and place task, where the EE alternates
between two operating points as shown in Fig. \ref{fig_2RRR_PKM}. The EE
trajectory follows a $\sin ^{2}$ time profile of the jerk. A quantization of
the joint angle encoder of 1/2000 is assumed, and white noise with amplitude
of $3\cdot 10^{-4}$m is added to model sensor noise. Fig.~\ref%
{fig_flatness_based_control} shows the EE tracking error and the actuation
torques of the flatness-based controller. When exact measurement and
feedback is assumed, perfect tracking is achieved. Experimental results are
reported in \cite{GAMM2023}. 
\begin{figure}[tbh]
\centering\includegraphics[width=0.49\linewidth]{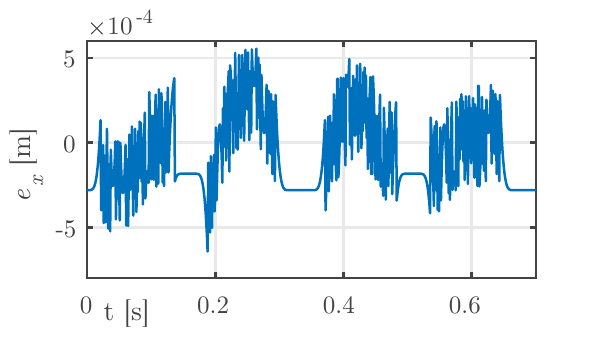}~ %
\includegraphics[width=0.49\linewidth]{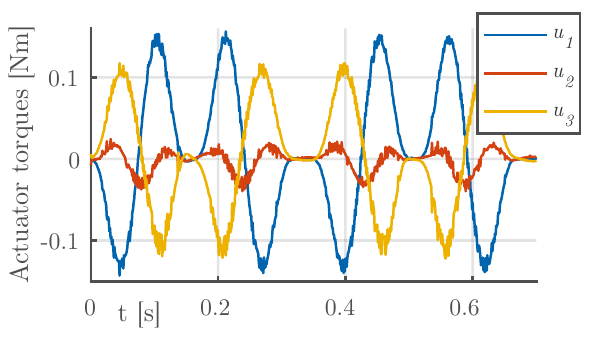}\vspace{-1ex}
\caption{Results of flatness-based control in task space. Tracking error $%
e_{x},e_{y}$, and actuator torques $u_{1},u_{2},u_{3},$}
\label{fig_flatness_based_control}
\end{figure}

\section{Conclusion and Outlook%
\label{secConclusion}%
}

A recursive algorithm for efficient computation of first and second time
derivatives of the inverse dynamics solution of non-redundant PKM with
simple limbs (each limb a serial chain) is presented, which also solves the
higher-order inverse kinematics problem. The Lie group framework is employed
to this end as it offers compact coordinate invariant expressions. The
formulation separates the overall kinematics of the PKM and the dynamics of
the individual limbs. It is exploits inverse dynamics formulations for
serial robots and a dedicated PKM modeling approach. This algorithm has $%
O\left( L\cdot n_{l}\right) $ complexity, except for the inversion of the
forward kinematics Jacobian. The geometric inverse kinematics was not
addressed as this depends on the particular PKM. A Newton-Raphson can always
be used, however \cite{Merlet2006,MuellerRAS2022}.

The proposed formulation is a solution to a key challenge in trajectory
tracking control of SEA-PKM. The modularity of the approach allows for
further efficiency improvements, in particular as it immediately allows for
parallelization. The exact computational effort of the recursive inverse
dynamics algorithm will be investigated in future research. The complexity
will be compared when using the closed form expression for the inverse
dynamics of the limbs \cite{MUBO2021}. Future work includes extension of the
second-order inverse dynamics algorithms to PKM with complex limbs (limbs
with closed loops). Further possible extension of this work includes dealing
with series-parallel hybrid robots, e.g. in humanoids with series elastic
actuators~\cite{2022_boukheddimi, 2015_valkyrie}. In addition to the Matlab
implementation, a C++ implementation of the generic version of the algorithm
is planned in Hybrid Robot Dynamics (HyRoDyn)~\cite{2020_hyrodyn} software
framework.

\section*{Acknowledgement}

Support by LCM K2 Center for Symbiotic Mechatronics within the Austrian
COMET-K2 program is acknowledged.

\section*{Appendix: Lie-Group Modeling of Kinematics}

The kinematics is described using the Lie group formulation introduced in 
\cite{MUBOScrews1,MMT2022}, which is a variant of that reported in \cite%
{LynchPark2017}. For notation and background of the Lie group modeling,
refer to \cite{Murray,Selig2005,LynchPark2017}. Denote with $\mathbf{C}%
_{i\left( l\right) }\in SE\left( 3\right) $ the configuration (pose) of body 
$i=1,\ldots ,N_{l}$ of limb $l$ relative to a spatial inertial frame, which
is represented as $4\times 4$ homogenous transformation matrices. It is
determined by the product of exponentials (omitting limb index $\left(
l\right) $)%
\begin{equation}
\mathbf{C}_{i}\left( 
\mathbold{\vartheta}%
\right) =\mathbf{B}_{1}\exp (\mathbf{X}_{1}\vartheta _{1})\cdot \ldots \cdot 
\mathbf{B}_{i}\exp (\mathbf{X}_{i}\vartheta _{i})  \label{POEX}
\end{equation}%
with the screw coordinate vector $\mathbf{X}_{i}\in {\mathbb{R}}^{6}$ of
joint $i$ represented in the reference frame at body $i$, and $\mathbf{B}%
_{i}\in SE\left( 3\right) $ is the configuration of body $i$ relative to
body $i-1$ in the zero reference $%
\mathbold{\vartheta}%
_{\left( l\right) }=\mathbf{0}$. The platform is the terminal body of each
limb $l$ with $N_{l}$ bodies, and its configuration is $\mathbf{C}_{\mathrm{p%
}\left( l\right) }\equiv \mathbf{C}_{N_{l}\left( l\right) }$. The
configuration of body $i$ relative to the platform is $\mathbf{C}_{\mathrm{p}%
,i}:=\mathbf{C}_{\mathrm{p}}^{-1}\mathbf{C}_{i}$. The instantaneous screw
coordinate vector of joint $i=1,\ldots ,N_{l}$ expressed in platform frame
is $\mathbf{J}_{\mathrm{p}\left( l\right) ,i}=\mathbf{Ad}_{\mathbf{C}_{%
\mathrm{p},i}}\mathbf{X}_{i}$. This is the $i$-th column of the $6\times
N_{l}$ geometric forward kinematics Jacobian of limb $l$ in (\ref{FWKin}) 
\begin{equation}
\mathbf{J}_{\mathrm{p}\left( l\right) }=\left( 
\begin{array}{cccc}
\mathbf{Ad}_{\mathbf{C}_{\mathrm{p},1}}\mathbf{X}_{1} & \mathbf{Ad}_{\mathbf{%
C}_{\mathrm{p},2}}\mathbf{X}_{2} & \cdots  & \mathbf{X}_{N_{l}}%
\end{array}%
\right) .  \label{Jp}
\end{equation}%
Therein, $\mathbf{Ad}_{\mathbf{C}_{i,j}\text{ }}$is the $6\times 6$ matrix
transforming twist/screw coordinates when expressed in the frame on body $j$
to those when expressed in the frame on body $i$. It is also called
'composite body transformation operator' \cite{Jain1991} or 'twist
propagation matrix' \cite{SahaSchiehlen2001}. Denote with $\mathbf{ad}$ the $%
6\times 6$ 'spatial cross product' matrix \cite{Jain1991,Featherstone2008}
describing the Lie bracket.

\begin{lemma}
The time derivatives of the geometric Jacobian admit the closed form
expressions (omitting limb index $\left( l\right) $)%
\begin{align}
\dot{\mathbf{J}}_{\mathrm{p},i}& =-\mathbf{ad}_{\Delta \mathbf{V}_{\mathrm{p}%
,i}}\mathbf{J}_{\mathrm{p},i}  \label{Jpd} \\
\ddot{\mathbf{J}}_{\mathrm{p},i}& =%
\big%
(\mathbf{ad}_{\Delta \mathbf{V}_{\mathrm{p},i}}^{2}-\mathbf{ad}_{\Delta \dot{%
\mathbf{V}}_{\mathrm{p},i}}%
\big%
)\mathbf{J}_{\mathrm{p},i}  \label{Jpdd} \\
\dddot{\mathbf{J}}_{\mathrm{p},i}& =%
\big%
(-\mathbf{ad}_{\Delta \mathbf{V}_{\mathrm{p},i}}^{3}+2\mathbf{ad}_{\Delta 
\dot{\mathbf{V}}_{\mathrm{p},i}}\mathbf{ad}_{\Delta \mathbf{V}_{\mathrm{p}%
,i}}  \label{Jpddd} \\
& \ \ \ \ \ +\mathbf{ad}_{\mathbf{ad}_{\Delta \mathbf{V}_{\mathrm{p}%
,i}}\Delta \dot{\mathbf{V}}_{\mathrm{p},i}}-\mathbf{ad}_{\Delta \ddot{%
\mathbf{V}}_{\mathrm{p},i}}%
\big%
)\mathbf{J}_{\mathrm{p},i}  \notag
\end{align}%
with the twist of body $i$ relative to the platform, represented in the
platform frame, and its derivatives given as%
\begin{align}
\Delta \mathbf{V}_{\mathrm{p},i}& =\mathbf{V}_{\mathrm{p}}-\mathbf{Ad}_{%
\mathbf{C}_{\mathrm{p},i}}\mathbf{V}_{i}  \label{DV} \\
\Delta \dot{\mathbf{V}}_{\mathrm{p},i}& =\dot{\mathbf{V}}_{\mathrm{p}}-%
\mathbf{Ad}_{\mathbf{C}_{\mathrm{p},i}}\dot{\mathbf{V}}_{i}+\mathbf{ad}%
_{\Delta \mathbf{V}_{\mathrm{p},i}}\mathbf{V}_{\mathrm{p}}  \label{DVd} \\
\Delta \ddot{\mathbf{V}}_{\mathrm{p},i}& =\ddot{\mathbf{V}}_{\mathrm{p}}-%
\mathbf{Ad}_{\mathbf{C}_{\mathrm{p},i}}\ddot{\mathbf{V}}_{i}  \label{DVdd} \\
& \ \ \ \ +\mathbf{ad}_{\Delta \dot{\mathbf{V}}_{\mathrm{p},i}}\left( 
\mathbf{V}_{\mathrm{p}}-\Delta \mathbf{V}_{\mathrm{p},i}\right) -\mathbf{ad}%
_{\Delta \mathbf{V}_{\mathrm{p},i}}^{2}\mathbf{V}_{\mathrm{p}}.  \notag
\end{align}
\end{lemma}

\begin{proof}
\parindent0pt%
The relation $\dot{\mathbf{Ad}}_{\mathbf{C}_{\mathrm{p},i}}=-\mathbf{ad}%
_{\Delta \mathbf{V}_{\mathrm{p},i}}\mathbf{Ad}_{\mathbf{C}_{\mathrm{p},i}}$ 
\cite{MUBOScrews1} applied to $\mathbf{J}_{\mathrm{p}\left( l\right) ,i}=%
\mathbf{Ad}_{\mathbf{C}_{\mathrm{p},i}}\mathbf{X}_{i}$, with constant $%
\mathbf{X}_{i}$, yields (\ref{Jpd}). Repeated application yields (\ref{Jpd})
and (\ref{Jpdd}). With (\ref{DV}) follows $\Delta \dot{\mathbf{V}}_{\mathrm{p%
},i}=\dot{\mathbf{V}}_{\mathrm{p}}-\dot{\mathbf{Ad}}_{\mathbf{C}_{\mathrm{p}%
,i}}\mathbf{V}_{i}-\mathbf{Ad}_{\mathbf{C}_{\mathrm{p},i}}\dot{\mathbf{V}}%
_{i}=\dot{\mathbf{V}}_{\mathrm{p}}-\mathbf{Ad}_{\mathbf{C}_{\mathrm{p},i}}%
\dot{\mathbf{V}}_{i}-\mathbf{ad}_{\mathbf{V}_{\mathrm{p}}}\left( \mathbf{V}_{%
\mathrm{p}}-\mathbf{Ad}_{\mathbf{C}_{\mathrm{p},i}}\mathbf{V}_{i}\right) $
and hence (\ref{DVd}). The second derivative $\Delta \ddot{\mathbf{V}}_{%
\mathrm{p},i}=\ddot{\mathbf{V}}_{\mathrm{p}}-\mathbf{Ad}_{\mathbf{C}_{%
\mathrm{p},i}}\ddot{\mathbf{V}}_{i}-\mathbf{ad}_{\Delta \mathbf{V}_{\mathrm{p%
},i}}\mathbf{Ad}_{\mathbf{C}_{\mathrm{p},i}}\dot{\mathbf{V}}_{i}+\mathbf{ad}%
_{\Delta \dot{\mathbf{V}}_{\mathrm{p},i}}\mathbf{V}_{\mathrm{p}}+\mathbf{ad}%
_{\Delta \mathbf{V}_{\mathrm{p},i}}\dot{\mathbf{V}}_{\mathrm{p}}$, can be
reformulated to (\ref{DVdd}).
\end{proof}

\section*{List of main Symbols}

\begin{itemize}
\item ${\mathbold{\vartheta}\in {\mathbb{R}}^{N}}$ - joint coordinate vector
of PKM mechanism

\item ${\mathbold{\vartheta}}_{\mathrm{a}}\in {\mathbb{R}}^{n_{\mathrm{a}}}$
- joint variable vector of SEA-actuated joints

\item $\mathbf{q}_{\mathrm{m}}\in {\mathbb{R}}^{n_{\mathrm{a}}}$ - vector of
joint variables of SEA-drive units

\item $\mathbf{u}\in {\mathbb{R}}^{n_{\mathrm{a}}}$ - drive torques/forces
at SEA-actuated joints

\item $\bar{\mathbf{Q}}_{\left( l\right) }\in {\mathbb{R}}^{n_{l}}$ - vector
of generalized forces of limb $l$

\item $\mathbf{J}_{\mathrm{IK}}$ - $n_{\mathrm{a}}\times \delta _{\mathrm{p}%
} $ inverse kinematics Jacobian PKM

\item $\mathbf{F}_{\left( l\right) }$ - $n_{l}\times \delta _{\mathrm{p}}$
inverse kinematics Jacobian of limb $l$

\item $\mathbf{C}_{i}\in SE\left( 3\right) $ - absolute configuration of
body $i$, i.e. relative to the inertial frame (IFR)

\item $\mathbf{B}_{i}\in SE\left( 3\right) $ - reference configuration of
body $i$ relative to its predecessor

\item $\mathbf{X}_{i}\in {\mathbb{R}}^{6}$ - screw coordinate vector (ray
coordinates) of joint $i$ represented in the reference frame on body $i$

\item $\mathbf{V}_{i}\in {\mathbb{R}}^{6}$ - twist coordinate vector (ray
coordinates) of body $i$ represented in body-frame $i$

\item $\mathbf{V}_{\mathrm{t}}\in {\mathbb{R}}^{\delta _{\mathrm{p}}}$ -
task space velocity vector expressed in platform frame

\item $\mathbf{W}_{i}\in {\mathbb{R}}^{6}$ - wrench coordinate vector (axis
coordinates) represented in body-frame $i$

\item $\mathbf{Ad}_{\mathbf{C}_{i,j}}$- $6\times 6$ twist transformation
matrix, body $j$ to $i$

\item $\mathbf{M}_{i}$ - $6\times 6$ body-fixed mass matrix of body $i$
\end{itemize}

$n_{\mathrm{a}}$ - \# actuated joints, $N$ - total \# joint variables, $N_{l}
$ - \# joint variables of limb $l$ when connected to platform, $n_{l}$ - \#
tree-joint variables of limb $l$ (when disconnected from platform). $\delta
_{\mathrm{p}}$ - DOF of PKM platform. $\delta $ - DOF of mechanism.

\bibliographystyle{IEEEtran}
\bibliography{ICRA2023}

\end{document}